\documentclass[11pt,a4paper]{article}
\usepackage[hyperref]{acl2020}
\usepackage{times}
\usepackage{latexsym}

\usepackage{microtype}
\usepackage{amsmath}
\usepackage{amsfonts}
\usepackage{mathrsfs}
\usepackage{cleveref}
\usepackage[switch]{lineno}
\usepackage{xcolor}
\usepackage{graphicx}
\usepackage{multirow}

\aclfinalcopy

\title{A Simple and Efficient Multi-Task Learning Approach for Conditioned Dialogue Generation}

  \author{Yan Zeng \\
  DIRO, Université de Montréal \\
  \texttt{yan.zeng@umontreal.ca} \\\And
  Jian-Yun Nie \\
  DIRO, Université de Montréal \\
  \texttt{nie@iro.umontreal.ca} \\}

\date{}

\begin{document}
\maketitle

\begin{abstract}
Conditioned dialogue generation suffers from the scarcity of labeled responses. In this work, we exploit labeled non-dialogue text data related to the condition, which are much easier to collect. We propose a multi-task learning approach to leverage both labeled dialogue and text data. The 3 tasks jointly optimize the same pre-trained Transformer -- conditioned dialogue generation task on the labeled dialogue data, conditioned language encoding task and conditioned language generation task on the labeled text data. Experimental results show that our approach outperforms the state-of-the-art models by leveraging the labeled texts, and it also obtains larger improvement in performance comparing to the previous methods to leverage text data.
\end{abstract}

\section{Introduction}
General conversational models pre-trained on large text data \cite{radford2018improving, devlin2018bert} or human-to-human conversation data \cite{zhang2019dialogpt, bao2019plato} have shown excellent performance in generating fluent and diverse responses. In addition to general conversation, we are more and more faced with the problem of conditioned conversation that tunes the dialogue toward a specific style or domain. 
For example, we might specify a condition as the vocabulary frequently used by a person and ask the system to mimic the speaking style of the person, or a topic-related vocabulary and ask the chatbot to discuss the given topic. 

Conditioned response generation has been extensively explored using RNN-based sequence-to-sequence models, under different conditions, e.g. persona \cite{li2016persona}, topic \cite{xing2017topic}, emotion \cite{zhou2018emotional}, situations \cite{sato2017modeling}, and so on. However, only a few existing studies considered using pre-training based models \cite{zheng2019pre, lin2019caire}. The basic idea in these previous works is to utilize a parametric vector to represent a condition and then use it in the decoder for conditioned generation. However, the key issue in conditioned dialogue generation is the availability of labeled responses \cite{zhou2018mojitalk}, and pre-training on unlabeled text or dialogue data does not help much. Therefore, the motivation of this work is to leverage labeled text (non-dialogue) data that are much easier to collect than labeled dialogue data as supplement. These data can be, for example, texts written by the same person (for a persona condition), within the same topic domain (for a topic condition), etc. The idea is inspired by response style transfer \cite{luan2017multi, niu2018polite}, which uses a text corpus to learn a style and then transfer the style to dialogue. Based on their success, we assume that the labeled text data can contribute to create better representations of conditions and better utilization of conditions in natural language generation. 

In this work, we propose a multi-task learning approach to leverage both labeled dialogue and text data. We use 3 tasks to jointly optimize the same pre-trained Transformer -- conditioned dialogue generation task on the labeled dialogue data, conditioned language encoding task and conditioned language generation task on the labeled text data. Our assumption is that the two other tasks can help in our final goal of conditioned dialogue generation: conditioned language generation is the base of conditioned response generation, and conditioned language encoding using bi-directional attention can efficiently encode condition-related expressions and lead to better condition representations. We apply different input representations, self-attention masks, and random mask strategies to differentiate the 3 tasks. Regardless of these differences, the training objectives of these tasks are essentially the same, i.e. masked language modeling, 
and thus we can mix up 2 types of data / 3 tasks  in one training batch, which prevents us from having the catastrophic forgetting problem \cite{phang2018sentence}. 

To efficiently leverage labeled data, first, our approach incorporates all types of data within the same framework, avoiding introducing ad hoc model components which are usually needed in some response style transfer methods in order to leverage extra texts. Second, we propose \emph{TF-IDF based masking} which selects more condition-related tokens to mask, so that the model can exploit the labeled text data more for 
condition-related expressions rather than the general language features already captured by the pre-trained models. 
Third, for conditioned generation, we propose a \emph{non-parametric attention-based gating mechanism}, which chooses between generating a general word (necessary for general function words) or a condition-related word at each position. We expect it to be more efficient than a parametric gating. Experimental results show that these approaches all bring improvements. 

Our approach is generalizable. In spite of many different labels, a condition essentially specifies some preferences on words, phrases, and sentence structures in the generated responses. Thus, a general approach can be instantiated to a specific case as long as the corresponding labeled dialogue data are available. We will run experiments with two instantiated models for persona- and topic-related dialogue. Additionally, we will empirically show that our approach is robust and can even work with condition labels predicted by a classification model, e.g. LDA for topic labels.

The contributions in this work are as follows \footnote{Code is available at \url{https://github.com/zengyan-97/MultiT-C-Dialog}.}: 

\begin{itemize}
\item We propose a simple and efficient multi-task learning approach based on pre-trained Transformer that leverages different labeled data, i.e., dialogue and text, for conditioned response generation.  

\item The experiments under two different conditions -- persona- and topic-based dialogue, show that our approach outperforms the state-of-the-art models by leveraging labeled texts even when the labels are predicted by a model.

\item Our approach obtains larger improvement in performance comparing to the existing methods to leverage text data, based on extra auto-encoder or  sequential fine-tuning. 

\end{itemize}

\section{Related Work}
\subsection{Conditioned Dialogue Generation}
\label{sec:related_c}
We categorize the related existing works into 3 categories.  
(1) Response generation conditioned on latent variables, where no extra annotations of dialogues is required \cite{serban2017hierarchical, shen2018improving, gu2018dialogwae, chen2019generating, gao2019generating, bao2020plato}. 
(2) Loosely-conditioned response generation, where a label designating the type of the response is required. For example, persona labels \cite{li2016persona} designate the speaking styles of the responses, and topic labels \cite{xing2017topic, dziri2019augmenting} or emotion labels \cite{li2017dailydialog, zhou2018emotional, rashkin2019towards} specify topic-related or emotion-related vocabularies. These studies usually utilize a parametric vector to encode a label, which is then used in the decoder to guide the generation. (3) Strictly-conditioned response generation, where extra knowledge is required to determine the content of the response, such as a persona profile \cite{zhang2018personalizing, urbanek2019learning}, a situation description \cite{rashkin2018towards, urbanek2019learning}, or a wikipedia paragraph \cite{galley2019grounded, dinan2018wizard}, which are used to ground the response. The ability to strictly-conditioned generation is important, but these dialogues only count for a small fraction of open-domain conversation \cite{zheng2019pre}. In many other cases, we are in the situation of loosely-conditioned dialogue. 
Furthermore, the state-of-the-art strictly-conditioned method \cite{wolf2019transfertransfo} can be easily added in other models as well \cite{shuster2019dialogue, madotto2020adapter}, which simply concatenates the extra knowledge with the dialogue history as the model input. 

In this work, we focus on loosely-conditioned response generation \footnote{Conditioned generation elsewhere in this work refers to loosely-conditioned generation.}. We will show that our approach is robust and can work with different types of labels including those predicted by a classification model, e.g. LDA for topic labels. Therefore, our method is compatible to generation conditioned on latent variables by borrowing power of a classification model. In this work, we do not touch on strictly-conditioned generation. However, this ability can be easily equipped as mentioned.

\subsection{Response Style Transfer}
Style transfer in dialogue aims to learn the style of a text corpus and then incorporate the style in dialogue generation. The transfer is usually between two styles, e.g. rude and polite, or adding a style to general dialogues. To leverage the text corpus, \citet{luan2017multi} jointly trains a seq2seq response generator and an extra auto-encoder, and \citet{niu2018polite} trains an extra style classifier first to guild the response generator using reinforcement learning. 

These works show that text data contain rich information about how to generate a specific type of texts, which inspire us to exploit the labeled text data in conditioned dialogue generation to alleviate the data scarcity issue. Style transfer is usually between two given styles. In contrast, conditioned dialogue generation could work with hundreds of condition labels simultaneously. As we will show in our experiments, the style transfer methods that utilize additional models, e.g. auto-encoder, to leverage text corpus are unscalable and inefficient for conditioned dialogue. In contrast, our approach that leverages labeled text data without using ad hoc models and makes a tighter integration of labeled text data with labeled dialogue data can more directly impact the conditioned dialogue generation.

\section{Method}
\begin{figure*}[t]
\centering
\includegraphics[height=2.8in]{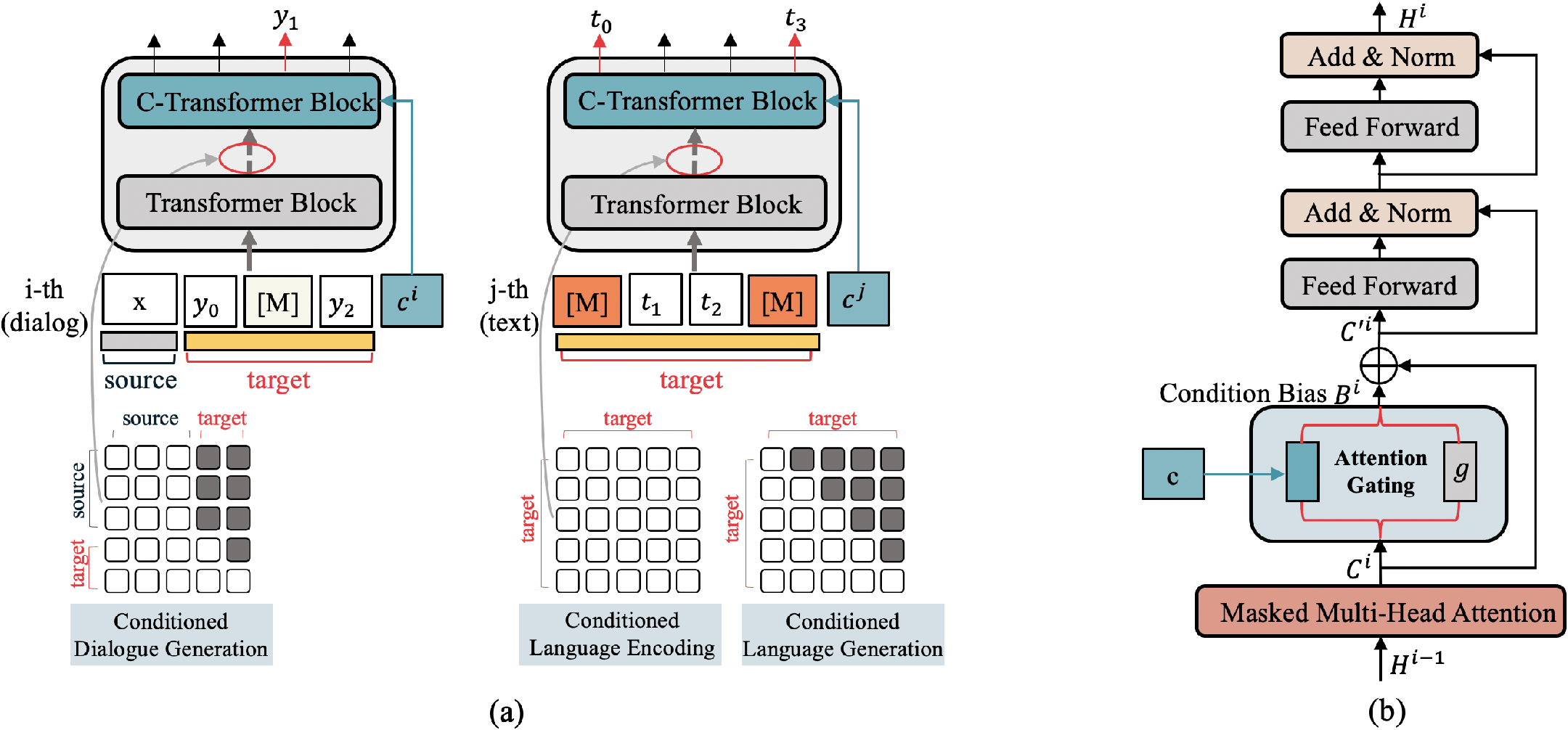}
\caption{(a) Overview of our multi-task learning approach. Labeled dialogue and text data are mixed, and they are processed using the same pre-trained Transformer with data/task-adaptive input representations, self-attention masks, and random mask strategies. (b) Detailed structures of a condition-aware transformer block, i.e. a C-Transformer Block.}
\label{Fig:input}
\end{figure*} 

We assume that we have two types of training data: a labeled dialogue corpus containing (dialogue history, condition, response) samples, and a labeled text corpus consisting of (condition, text) samples. Notice that the ``condition'' is any categorical label that indicates a type of responses or texts. Our goal is to generate a response $y$ that exhibits the desired characteristics of the type of responses given a dialogue history $x$ and a condition $c$: 
\begin{equation}
y = \mathop{\arg\max}_{y}P(y|x, c) \ \
\end{equation}

The Transformer in our work uses bi-directional attention on the source side to encode the dialogue history, and left-to-right attention on the target side to generate the response. Such a transformer can be initialized from BERT\cite{devlin2018bert}, Roberta\cite{liu2019roberta}, UniLM \cite{dong2019unified}, or the models pre-trained on large-scale unlabeled dialogue data e.g. PLATO \cite{bao2019plato} and Blender \cite{roller2020recipes}. In this work, we focus on efficiently leveraging labeled data, i.e. dialogue and text. Figure \ref{Fig:input} (Left) shows the overview of our approach.

\subsection{Masked Multi-Head Attention}
\label{sec:maskattn}
In this subsection, we introduce the basic components of Transformer. Masked multi-head attention is also applied in our condition-aware transformer block. The input representation $\mathbf{H}^0 \in \mathbb{R}^{n \times d_h}$, where $n$ is the input length and $d_h=768$ is the hidden dimension, is the sum of token embedding, position embedding, and type embedding at each position. We apply type embeddings to introduce a separation between source side and target side as shown in Figure \ref{Fig:input} (Left) in order to warrant different treatments in the model. Then, $\mathbf{H}^0$ is encoded into hidden representations of $i$-th layer $\mathbf{H}^i=[\mathbf{h}_1^i, ...,\mathbf{h}_{n}^i]$ using multi-layer transformer blocks: $\mathbf{H}^i={\rm Trans}^i(\mathbf{H}^{i-1}) \quad i \in [1,L]$. The core component of a transformer block is the masked multi-head attention, whose outputs, i.e. contextualized representations, $\mathbf{C}^i=[\mathbf{c}_1^i, ...,\mathbf{c}_{n}^i]$, are computed via: $\mathbf{C}^i = {\rm Concat}(\mathbf{head}_1, ..., \mathbf{head}_h)$. Specifically, 
\begin{equation}
\mathbf{head}_j = {\rm softmax}(\frac{\mathbf{Q}_j\mathbf{K}_j^{T}}{\sqrt{d_k}}+\mathbf{M})\mathbf{V}_j
\end{equation}
\noindent where $\mathbf{Q}_j, \mathbf{K}_j, \mathbf{V}_j \in \mathbb{R}^{n\times d_k}$ are obtained by transforming $\mathbf{H}^{i-1} \in \mathbb{R}^{n\times d_h}$ using $\mathbf{W}_{j}^{iQ}, \mathbf{W}_j^{iK}, \mathbf{W}_j^{iV} \\ \in \mathbb{R}^{d_h\times d_k}$ respectively. The self-attention mask matrix $\mathbf{M} \in \mathbb{R}^{n\times n}$ (with $\mathbf{M}_{ij} \in \{0, -\infty \}$) determines whether a position can attend to other positions:  $\mathbf{M}_{ij}=0$ allows the $i$-th position to attend to $j$-th position and $\mathbf{M}_{ij}=-\infty$ prevents from it.

Our approach jointly optimizes three tasks that apply different self-attention masks as shown in Figure \ref{Fig:input} (Left). For conditioned dialogue generation task, the self-attention mask allows bi-directional attention on the source side to fully encode dialogue history and left-to-right attention 
on the target side to generate conditioned responses. For the labeled text data, we randomly choose between conditioned language encoding and conditioned language generation task. The two tasks use bi-directional attention and left-to-right attention respectively. The language encoding objective, i.e. Masked Language Modeling (MLM), is used in BERT, which has shown stronger ability than the auto-regressive objective used in GPT \cite{devlin2018bert}. Therefore, we expect conditioned language encoding is more helpful to learn condition-related expressions (especially with the TF-IDF masking strategy which we will introduce) than the two generation tasks that employ the auto-regressive objective. 

\subsection{Condition-aware Transformer Block}
\label{sec:cbias}
In this subsection, we introduce position-wise condition bias that aims to determine how much condition information should be utilized to bias word generation probability at a position. The core component to calculate the bias is a \textbf{non-parametric attention-based gating mechanism} as shown in Figure \ref{Fig:input} (Right). Other gate mechanisms usually employ parametric linear layers to calculate weights. We assume a self-attention based method (non-parametric) could be more training-efficient, which is important since labeled data are usually limited. We will empirically confirm its effectiveness compared to other gating methods.

Specifically, given a training sample (x, c, y) or (c, text), the condition label $c$ is encoded using two sets of parameters: one parametric vector works as the key $\mathbf{k}^c \in \mathbb{R}^{d_h}$ and another one works as the value $\mathbf{v}^c \in \mathbb{R}^{d_h}$. Additionally, there is a general condition label $g$ with a parametric vector $\mathbf{k}^{g}$ as its key and a zero vector $\mathbf{v}^{g}$ as its value. The former corresponds to conditioned generation, while the latter to the general dialogue that generates words only based on dialogue history. At each position, the model determines an attention weight to each choice. More attention to $c$ means that the position is more tuned to the condition. More specifically, for each condition-aware transformer block as shown in Figure \ref{Fig:input}(Right), given  $\mathbf{C}^i= 
[\mathbf{c}_1^i, ..., \mathbf{c}_{n}^i]$ as queries, the condition biases $\mathbf{B}^i = [\mathbf{b}_1^i, ..., \mathbf{b}_{n}^i]$ are calculated by: 
\begin{equation}
\mathbf{B}^i = {\rm softmax}(\frac{\mathbf{C}^i \mathbf{K}_b^T}{\sqrt{d_k}}+\mathbf{M}_b)\mathbf{V}_b
\end{equation}
where $\mathbf{K}_b=[\mathbf{k}^c, \mathbf{k}^g]$ and $\mathbf{V}_b=[\mathbf{v}^c, \mathbf{v}^g]$. The calculation is non-parametric. We use the matrix $\mathbf{M}_b \in \mathbb{R}^{n\times 2}$ to prevent adding condition bias to positions on the source side because the condition only influences the target side (the labeled response or text).

\subsection{Objectives}

We jointly optimize three tasks: conditioned dialogue generation on labeled dialogue, conditioned language encoding and conditioned language generation on labeled text. As discussed in Section \ref{sec:maskattn}, conditioned language encoding is expected to be very helpful to learn condition-related expressions.   

A specific self-attention mask is required for each task, while the objectives of three tasks are essentially the same -- some tokens of the target side (labeled response or text) are randomly masked, and the final hidden vectors $H^L$ corresponding to the masked tokens are fed into an output softmax over the vocabulary to predict the expected tokens. Therefore, we can mix up 2 types of data (3 different tasks) in one training batch, and the loss is averaged in a batch. This thus prevents us from having the catastrophic forgetting problem \cite{phang2018sentence}. This problem is usually observed using a sequential fine-tuning process, i.e. first fine-tuning on labeled texts and then on conditioned dialogue data, which will erase the effect of the previous steps of training. 

When using labeled dialogue data, we want the model to learn to generate conditioned but more importantly coherent responses. Thus, we uniformly sample the tokens on the target side to mask. Differently, when exploiting labeled text data, we only want the model to generate condition-related expressions. Therefore, we introduce \textbf{TF-IDF Based Masking} for the labeled text data to speed up the learning process -- we sample tokens to mask according to their TF-IDF values counted on the entire corpus. We will empirically show its effectiveness.

\section{Experiments}

\subsection{Datasets}
We use two labeled dialogue datasets, and we created two smaller training sets (500K labeled texts and 250K labeled dialogues), which are summarized in Table \ref{tab:dataset}. We anticipate that when labeled dialogue data are limited, the benefit of leveraging labeled text data will be larger. 

\textbf{Persona Reddit} We filtered the Reddit data from 2015 to 2019 that is provided by a third party \footnote{ \url{https://files.pushshift.io/reddit/}}. Reddit data is a natural source of dialogue with multiple users -- a post may have multiple comments by different users. Following \citet{li2016persona}, we consider each user as a distinct persona. We extract (post, user, comment) tuples, where ``user'' is the label of the user who makes the ``comment''. We further filtered the data based on sentence length and users: sentences with more than 30 words or less than 4 words are removed, and we only keep comments from the 2000 most active users so that we can collect enough data for each user. As a result, 
each user has 1291 samples (comments) on average. To build the labeled text corpus, we collect extra posts or comments on Reddit from the same user that have no overlap with the dialogue data -- these extra texts are intended to reflect the general writing style of the user.

\textbf{Topic-related Dialogue} \citet{dziri2019augmenting} provides a high-quality 3-turns conversational dataset for topic aware response generation \footnote{\url{https://github.com/nouhadziri/THRED}}. Along with each (history, target) pair, there is a topic label and dozens of topic words that are predicted by LDA model. 
The dataset contains 9.2M samples, from which we sample 3M (history, topic, target) tuples as the labeled dialogue corpus. To construct the labeled text data, we sample other 3M tuples and only keep their (topic, target) parts. Note that the topic labels are generated by LDA, and thus it is difficult to obtain the labeled text data from other sources.

\begin{table}[h]
\small
\begin{tabular}{l|ll|ll}
\hline 
\hline
Dataset & \multicolumn{2}{c|}{Persona Reddit} & \multicolumn{2}{c}{Topic dialogue} \\
\hline 
Source of Labels  & \multicolumn{2}{c|}{Personal ID} & \multicolumn{2}{c}{LDA} \\
Number of Labels & \multicolumn{2}{c|}{2000} & \multicolumn{2}{c}{190} \\
\hline
Labeled Texts & 3M & 500K & 3M & 500K \\
dialogue Train & 3M & 250K & 3M & 250K \\
dialogue Valid & \multicolumn{2}{c|}{80K}& \multicolumn{2}{c}{80K} \\
dialogue Test & \multicolumn{2}{c|}{10K} & \multicolumn{2}{c}{10K} \\
\hline
\end{tabular}
\caption{\label{tab:dataset} Key characteristics of the two datasets.
}
\end{table}

\subsection{Baselines}
We choose two strong baselines specifically designed for personalized response generation and two others for topic-aware generation. 
Additionally, we choose some state-of-the-art pre-trained Transformers. 

\paragraph{Speaker Model}\cite{li2016persona} a seq2seq model using four LSTM layers. Given a user label, the decoder transforms it into a user embedding and use it to generate a personalized response. 

\paragraph{MT-Speaker} an approach jointly trains a Speaker Model and a conditioned auto-encoder with shared decoder parameters, which is adapted from a style transfer approach \cite{luan2017multi}. This approach also leverages the labeled text data. 

\paragraph{TA-Seq2Seq} \cite{xing2017topic} and \textbf{THRED} \cite{dziri2019augmenting} these models utilize topic words instead of topic labels predicted by the LDA model. TA-Seq2Seq leverages the topic information by a joint attention mechanism and a biased generation probability. THRED is built based on HRED and incorporates topic words via a hierarchical joint attention mechanism. 

\paragraph{C-Trans-ED} \cite{zheng2019pre} an encoder-decoder transformer framework initialized with GPT parameters. The decoder dynamically merges features from the dialogue history and the condition. This model is based on the code of ConvAI2 champion \cite{dinan2019second}.

\paragraph{C-Trans-Dec} a decoder-only transformer initialized with GPT-2 parameters, adapted from \citet{wolf2019transfertransfo}. We add a condition embedding to the input representation to enable conditioned generation. 

\paragraph{BERT} fine-tuning the pre-trained model  \cite{devlin2018bert} on the dialogue datasets. The encoder and decoder share the parameters. When encoding, the model uses bi-directional attention. When decoding, it uses left-to-right attention.

\subsection{Implementation Details}
We implement the speaker model and MT-Speaker model based on OpenNMT \footnote{\url{http://opennmt.net/}}. Other models are directly taken from the available open-source code. Hyper-parameters are set following the original papers. 
Since our baselines utilize GPT or BERT, we use BERT (base, uncased) to initialize our model for fair comparison. It is however possible to 
build our model upon more powerful pre-trained models such as Roberta\cite{liu2019roberta}. We do hyper-parameter search based on perplexity on the validation set for: the number of condition-aware transformer blocks in $\{2, 6, 12\}$,  the mix-up rate of labeled dialogues and texts in $\{$3:1, 1:1$\}$, and whether using conditioned language encoding task.  
We report experimental results with 2, 3:1, and using conditioned language encoding respectively. 
The warm-up proportion is set to 0.1. $25\%$ tokens of the target side are randomly masked. During decoding the beam size is 10, and we prevent duplicated bigrams. We fine-tune all the parameters end-to-end for four epochs on two P100 GPUs. With in total 6M training samples, each epoch takes twelve hours. The fine-tuning model only has  $(2C+1) \times d_h$ additional parameters, where $C$ is the number of different condition labels. Other details are given in Appendix \ref{app:details}.

\subsection{Evaluation}
\paragraph{Automatic Metrics} We choose some widely used metrics in the literature \footnote{We use an open-source evaluation tool: \url{https://github.com/Maluuba/nlg-eval}}: \textbf{BLEU} \cite{papineni2002bleu} with n=1,2,3; \textbf{ROUGE-L} -- longest common subsequence based statistics; \textbf{CIDEr} \cite{vedantam2015cider} utilizing TF-IDF weighting for each n-gram; and \textbf{Distinct} \cite{li2016diversity} indicating the proportion of unique n-grams (n=1,2) in the entire set of generated responses to evaluate response diversity. Two-sided t-test is used for statistical significance test. 

\paragraph{Response Appropriateness} Furthermore, we conduct manual evaluation on the best models according to the automatic metrics. 
We only manually evaluate the model performance on large-scale datasets\footnote{We did not manually evaluate the results with small datasets due to its high cost. However, we expect even larger difference when small data are used for training, as indicated by the automatic metrics.}. 
We ask human evaluators to rate a response in $\{0, 1, 2\}$. A score of 0 means that the response might have flaw in fluency and logic or be incoherent. Special cases are for example completely coping from the dialogue history as the output, and a bland response such as ``I don't know what you mean''. A score of 1 represents a coherent but generic response. 2 represents a coherent and informative response. We also do a pair-wise evaluation to compare two models and indicate which one is better. The evaluation is based on 200 random samples. Each generated response is rated by three annotators. The inter-rater annotation agreement in Cohen’s kappa \cite{cohen1960coefficient} is $0.441$ on average, which indicates moderate agreement. 

\paragraph{Condition Consistency} We observe that automatic metrics fail to evaluate condition consistency since BERT that does not consider conditions outperforms C-Trans-ED and C-Trans-Dec. Thus, we perform manual evaluation on the condition consistency. 
A generated response is rated in $\{0, 1, 2\}$. The scores 0, 1 and 2 mean respectively that the response is inconsistent to the condition, somehow related, and consistent. However, if the response has flaw in fluency or logic, it will get a score of 0. For Topic Dialogue, it is easy to measure whether a generated response is in the topic. However, for persona consistency, it is difficult for a human evaluator to know the speaking style of each user. Thus, before evaluation we first automatically determine those frequently used words by a user in responses and show them to the annotators to help their evaluations. 

\begin{table*}[t]
\centering
\small
\begin{tabular}{llllllllll}
\hline
\hline
\textbf{Model} & \textbf{BLEU-1} & \textbf{BLEU-2} & \textbf{BLEU-3} & \textbf{ROUGE-L} & \textbf{CIDEr} & \textbf{Dist-1} & \textbf{Dist-2}  & \textbf{avgLen}\\ 
\hline
Sp-Model & 10.539 (**) & 3.152 (**) & 1.396 (**) & 0.116 (**) & 0.056 (**) & 0.012 (**) & 0.044 (**) & 12.6 \\

MT-Speaker & 10.970 (**) & 3.488 (**) & 1.540 (**) & 0.118 (**) & 0.059 (**) & 0.009 (**) & 0.034 (**) & 12.7 \\

C-Trans-ED & 13.548 (*) & 3.881 (**) & 1.529 (**) & 0.113 (**) & 0.045 (**) & 0.005 (**) & 0.025 (**) & 18.7 \\

C-Trans-Dec & 12.964 (**) & 4.182 (**) & 1.781 (**) & 0.117 (**) & 0.060 (**) & 0.023 (**) & 0.097 (**) & 16.7 \\

BERT & 12.928 (*) & 4.405 (/) & 1.764 (**) & 0.119 (**) & 0.062 (**) & 0.014 (**) & 0.052 (**)  & 26.1 \\

\hline
Ours & \textbf{14.052} & \textbf{4.891} & 2.149 & \textbf{0.122} & 0.070 &  \textbf{0.024} & 0.098 & 23.3 \\

\quad Two-Step FT & 13.714 (/) & 4.870 (/) & \textbf{2.160} (/) & \textbf{0.122} (/) & \textbf{0.071} (/) & 0.023 (/) & 0.102 (*) & 25.0 \\

\quad w/o ctext & 13.015 (*) & 4.563 (/) & 1.956 (/) & 0.113 (**) & 0.061 (**) & 0.023 (/) & \textbf{0.106} (*)  & 25.7 \\
\quad w/o tfidf & 13.581 (*) & 4.705 (/) & 2.000 (/) & 0.118 (**) & 0.070 (/) & 0.023 (/) & 0.095 (*) & 24.0 \\

\hline
\hline

Sp-Model & 10.467 (**) & 3.039 (**) & 1.239 (**) & 0.116 (**) & 0.049 (**) & 0.007 (**) & 0.027 (**) & 12.3 \\ 
MT-Speaker & 10.286 (**) & 2.932 (**) & 1.174 (**) & 0.114 (**) & 0.047 (**) & 0.007 (**) & 0.030 (**) & 12.3 \\ 

C-Trans-ED & 10.968 (**) & 3.247 (**) & 1.295 (**) & 0.106 (**) & 0.040 (**) & 0.001 (**) & 0.006 (**) & 14.7 \\

C-Trans-Dec & 11.263 (**) & 3.390 (**) & 1.274 (**) & 0.106 (**) & 0.043 (**) & 0.020 (**) & 0.075 (**) & 16.2 \\

BERT & 12.766 (*) & 4.195 (*) & 1.805 (*) & 0.118 (/) & 0.063 (*) & 0.022 (/) & 0.071 (**) & 15.3 \\

\hline
Ours & \textbf{13.517} & \textbf{4.517} & \textbf{1.988} & \textbf{0.119} & \textbf{0.068} & 0.021 & 0.066 & 16.4 \\

\quad Two-Step FT & 10.125 (**) & 3.295 (**)  & 1.388 (**)  & 0.111 (**)  & 0.052 (**)  & 0.015 (**)  & 0.043 (**) & 12.7 \\   

\quad w/o ctext & 11.776 (**) & 3.821 (**) & 1.631 (**) & 0.115 (**) & 0.059 (**) & 0.020 (*) & 0.062 (**) & 14.4 \\

\quad w/o tfidf & 13.475 (/) & 4.409 (/) & 1.853 (/) & 0.118 (/) & 0.064 (*) & \textbf{0.023} (/) & \textbf{0.078} (*) & 16.7 \\

\hline

\end{tabular}
\caption{\label{tab:persona} Evaluation results on large-scale (upper half) and small-scale (lower half) Persona Reddit. Two-Step FT means using our model architecture but applying sequential fine-tuning. w/o ctext is without leveraging conditioned text data. w/o tf-idf means without applying TF-IDF based masking. * ($p < 0.05$) or ** ($p < 0.01$) show statistically significant differences with our model by two-sided t-test. }
\end{table*}

\begin{table*}[t]
\centering
\small
\begin{tabular}{llllllllll}
\hline
\hline
\textbf{Model} & \textbf{BLEU-1} & \textbf{BLEU-2} & \textbf{BLEU-3} & \textbf{ROUGE-L} & \textbf{CIDEr} & \textbf{Dist-1} & \textbf{Dist-2} & \textbf{avgLen}\\ 
\hline
TA-Seq2Seq & 10.197 (**) & 3.307 (**) & 1.602 (**) & 0.121 (**) & 0.098 (**)  & 0.016 (**) & 0.051 (**) & 9.7 \\
THRED & 9.061 (**) & 3.035 (**) & 1.468 (**) & 0.118 (**) & 0.098 (**) & 0.015 (**) & 0.048 (**) & 8.8 \\

C-Trans-ED & 13.990 (**) & 5.359 (**) & 2.689 (**) & 0.131 (**) & 0.147 (**) & 0.055 (**) & 0.222 (**) & 12.5 \\

C-Trans-Dec & 14.544 (**) & 5.475 (**) & 2.669 (**) & 0.136 (**) & 0.154 (**) & 0.046 (**) & 0.177 (**) & 13.2 \\

BERT & 15.287 (/) & 6.243 (/) & 3.283 (/) & 0.141 (/) & 0.168 (**) & 0.057 (**) & 0.227 (**) & 12.5 \\

\hline
Ours & 15.639 & \textbf{6.484} & \textbf{3.455} & 0.140 & 0.185 & 0.060 & 0.243 & 13.0 \\ 

\quad Two-Step FT & \textbf{15.926} (/) & 6.431 (/) & 3.376 (/) & \textbf{0.143} (/) & 0.185 (/) & 0.059 (*) & 0.239 (*) & 13.1 \\

\quad w/o ctext  & 15.491 (*) & 6.397 (/) & 3.399 (/) & 0.142 (/) & \textbf{0.190} (*) & \textbf{0.063} (*) & \textbf{0.262} (**) & 12.8 \\

\quad w/o tfidf & 15.393 (/) & 6.302 (/) & 3.351 (/) & 0.139 (/) & 0.185 (/) & 0.059 (**) & 0.230 (**) & 13.1 \\

\hline
\hline

C-Trans-ED &  13.874 (**) & 5.145 (**) & 2.503 (*) & 0.124 (**) & 0.124 (**) & 0.039 (**) & 0.150 (**) & 13.1 \\

C-Trans-Dec & \textbf{14.899} (/) & 5.648 (/) & 2.690 (/) & 0.133 (**) & 0.150 (/) & 0.043 (*) & 0.176 (*) & 15.2 \\

BERT & 14.457 (/) & 5.583 (/) & 2.802 (/) & 0.135 (**) & 0.136 (**) & 0.037 (**) & 0.133 (**) & 12.4 \\
\hline
Ours & 14.587 & \textbf{5.747} & \textbf{2.894} & \textbf{0.139} & \textbf{0.152} & \textbf{0.050} & \textbf{0.186} & 12.0 \\

\quad Two-Step FT & 13.941 (**) & 5.463 (/) & 2.765 (/) & 0.136 (*) & 0.140 (**) & 0.045 (**) & 0.169 (**) & 11.7 \\   

\quad w/o ctext & 13.211 (**) & 5.179 (**) & 2.655 (/) & 0.137 (/) & 0.142 (**) & 0.046 (**) & 0.163 (**) & 10.8 \\

\quad w/o tfidf & 13.964 (**) & 5.485 (**) & 2.809 (/) & 0.135 (**) & 0.145 (*) & 0.048 (*) & 0.178 (**) & 11.8 \\

\hline 
\end{tabular}
\caption{\label{tab:topic} Evaluation results on large-scale and small-scale Topic Dialogue. Topic labels are predicted by LDA.}
\end{table*}

\begin{table*}[h]
\centering
\footnotesize
\begin{tabular}{l|llll|llll}
\hline
\hline
\multirow{3}{*}{\textbf{Model}} & \multicolumn{4}{c|}{\textbf{Persona}} & \multicolumn{4}{c}{\textbf{Topic}} \\

& \multicolumn{2}{c}{Appropriateness} & \multicolumn{2}{c|}{Consistency} &
\multicolumn{2}{c}{Appropriateness} & \multicolumn{2}{c}{Consistency} \\
 & Score & Pair-wise & Score & Pair-wise  & Score & Pair-wise & Score & Pair-wise  \\
\hline
C-Trans-Dec & 0.96 & (28\%, 39\%) & 0.85  & (20\%, 39\%)  & 0.77 & (26\%, 34\%) & 0.71 & (21\%, 31\%) \\
BERT & 0.77 & (11\%, 40\%) & 0.78  & (22\%, 43\%)  & 0.55 & (17\%, 40\%) & 0.46 & (16\%, 40\%) \\
Ours & \textbf{1.15} & - & \textbf{1.24}  &  & \textbf{0.83} & - & \textbf{0.80} & - \\
\quad w/o ctext & 0.91 & (26\%, 39\%) & 0.90 & (23\%, 38\%) & 0.73 & (27\%, 35\%) & 0.72 & (23\%, 30\%) \\

\hline
\end{tabular}
\caption{\label{tab:humaneval} Human evaluation of generated responses on appropriateness and condition consistency. Pair-wise comparisons show the wining percentages of (baseline, ours). }
\end{table*}

\subsection{Analysis}
Table \ref{tab:persona} and \ref{tab:topic} gives automatic evaluation results, and Table \ref{tab:humaneval} gives human evaluation results. Appendix \ref{app:cases} shows some generated responses. The results can be summarized as follow: 

\paragraph{BERT vs. Trans-ED \& Trans-Dec}
C-Trans-Dec has a clear advantage over C-Trans-ED in almost all automatic metrics, which can also be observed in their generated responses. Fine-tuning BERT without considering conditions outperforms C-Trans-Dec on most similarity metrics such as BLEU. We explain this by the fact that bi-directional attention could enable a model to better encode dialogue history, and thus to generate responses more similar to the ground truth. The ablation model using w/o ctext is fine-tuning C-BERT (with our condition-aware transformer blocks) on labeled dialogue data. The performance of w/o ctext is similar to C-Trans-Dec's, with a slight advantage in condition consistency and small disadvantage in response appropriateness. These results show that our approach is built upon a strong base model. As mentioned, other pre-trained models can also be used.

\paragraph{With Condition}
When large Persona Dialogue is available, w/o ctext (i.e. C-BERT) outperforms BERT in almost all automatic metrics. However, we observe that when only small-scale labeled dialogue data are available, all three conditioned models perform worse than BERT. 
This shows that the model cannot learn the condition-related features well from the limited labeled dialogue data. Thus, it is important to leverage the labeled texts that are easier to collect, and the results on small-scale Persona Reddit show that our multi-task learning approach significantly outperforms BERT on similarity metrics such as BLEU and CIDEr. 

For Topic Dialogue, the labels are given by LDA model. LDA is an unsupervised method and the predicted condition labels can be very noisy. Nevertheless, similarly, with large data C-BERT outperforms BERT in all metrics, but when only small-scale labeled dialogue data are available, C-BERT performs worse than BERT in terms of BLEU. The result again shows the importance of exploiting labeled texts, and our approach is the best on small-scale Topic Dialogue. 

\paragraph{Leveraging Labeled Texts} 

In general, our approach significantly outperforms all baselines and w/o ctext that do not exploit labeled text data, either with large-scale or small-scale data. With small-scale data, our approach outperforms BERT while w/o ctext itself cannot achieve this, which shows that conditioned dialogue generation can be helped by extra labeled text data. On Topic Dialogue, with such noisy labels, our model leveraging the labeled texts still produces the best performance, which confirms the robustness of our multi-task learning approach to work with different types of labels. The human evaluation on appropriateness and condition consistency further confirms the effectiveness of our approach.  

\begin{figure}[t]
\centering
\includegraphics[height=2.2in]{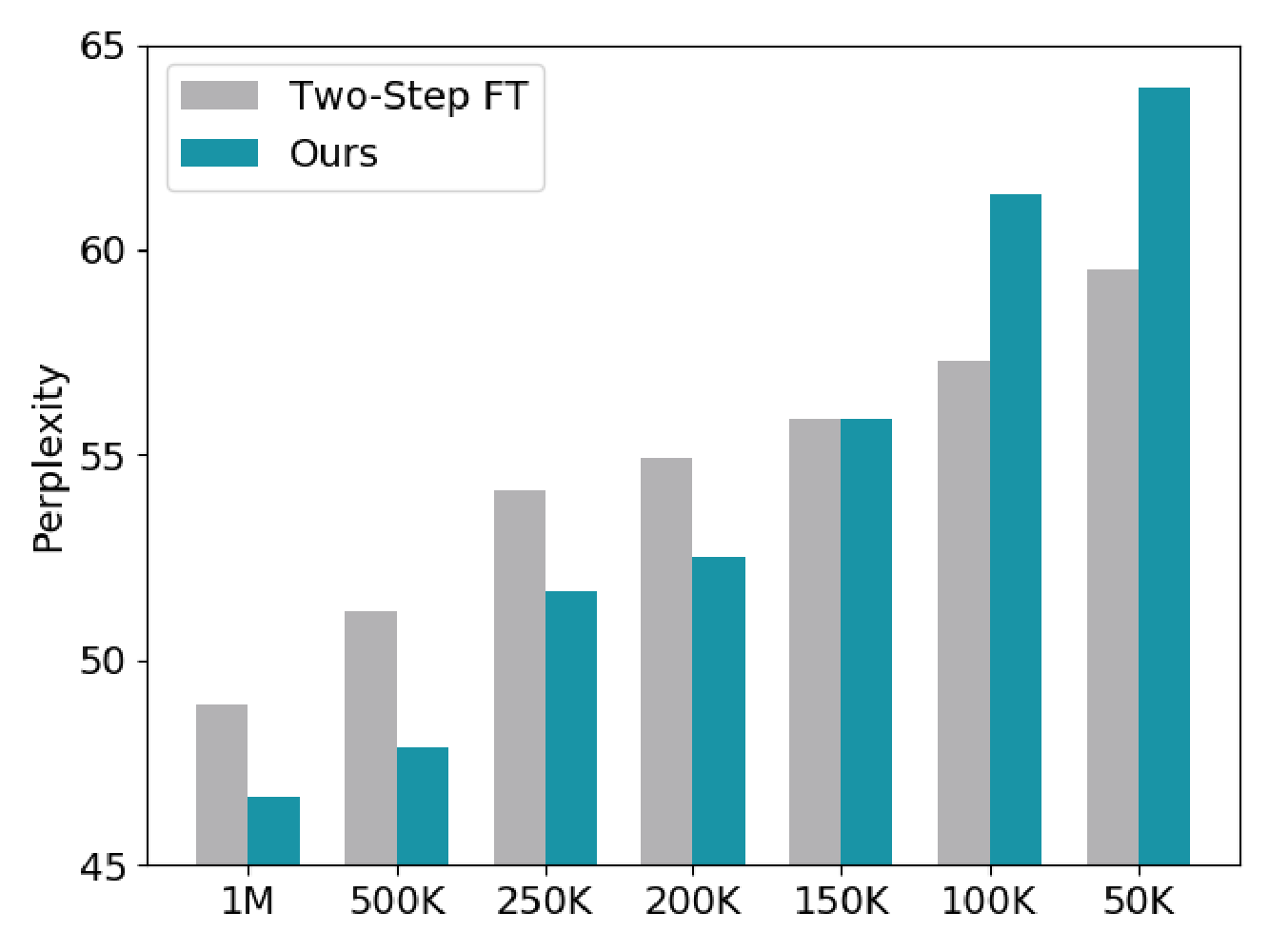}
\caption{Performance comparison between sequential fine-tuning and our approach given 1M labeled text data and different size of labeled dialogue data. }
\label{Fig:ppl}
\end{figure} 

Not all methods utilizing extra labeled text can obtain such performance improvement as we did. MT-Speaker that employs an extra auto-encoder does not gain much improvement over Sp-Model. This result shows  that using additional model components to leverage labeled texts is inefficient for conditioned dialogue generation. Furthermore, Two-Step FT that first fine-tuning on labeled texts and then on labeled dialogue data does not always produce good performance. It achieves comparable performance to our approach on large-scale datasets, but on small-scale datasets it can even perform worse than w/o ctext (Table \ref{tab:persona}). This result shows that the size of labeled text data is a considerable factor. With small-scale labeled text data, it is better to avoid sequential fine-tuning that has the catastrophic forgetting problem \cite{phang2018sentence}. Additionally, we investigate how the ratio of the size of labeled text data to the size of dialogue data influence model performance. As shown in Figure \ref{Fig:ppl}, given 1M labeled text data, when the ratio is less than $6.7$, our approach performs better than Two-Step FT. However, when labeled text corpus is much larger than dialogue corpus, sequential fine-tuning is better.

\begin{table}[t]
\centering
\small
\begin{tabular}{llllllllll}
\hline
\hline
\textbf{Model} & \textbf{BLEU-1} & \textbf{BLEU-2} & \textbf{Dist-2}\\ 
\hline

Single Gate & 13.880 (*) & 4.853 (/) & 0.090 (**)  \\
Double Gates & 13.988 (*) & 4.889 (/) & 0.094 (*) \\
Attn. Routing & \textbf{14.052} & \textbf{4.891} & \textbf{0.098}\\

\hline
\hline

Single Gate & 11.703 (**) & 3.891 (**) & 0.090 (**) \\
Double Gates & 11.336 (**) & 3.698 (**) & \textbf{0.091} (**) \\
Attn. Gating & \textbf{13.517} & \textbf{4.517} & 0.066 \\

\hline

\end{tabular}
\caption{\label{tab:gate} Comparison of gating mechanisms on large-scale and small-scale Persona Reddit.}
\end{table}

\paragraph{TF-IDF Masking and Attention Gating} 
We assumed that the general language features have already been captured by the pre-trained models. Thus, to better utilize labeled text data, we mask more condition-related words using TF-IDF based masking. Our ablation study confirms that TF-IDF masking brings improvement in almost all automatic metrics although the improvement might not always be statistically significant.

Our attention gating is a non-parametric gating mechanism to fuse the condition into the decoder. We expected it to be efficient, which is particularly important when labeled data are limited. Here, we compare it with two common parametric gating mechanisms: 1) setting a single gate on $\mathbf{C}^i$ to get a weight; 2) setting gates on both $\mathbf{C}^i$ and $\mathbf{v}^c$ to get two weights. Then, we combine the weighted $\mathbf{C}^i$ and $\mathbf{v}^c$ to get $\mathbf{C}^{'i}$ as in our attention gating. Experimental results in Table \ref{tab:gate} confirm that our method is more efficient. When only small-scale labeled data are available, the model with attention gating generates responses that are significantly more similar to the ground-truth.

\section*{Conclusion} 
In this paper, we examined the data scarcity issue of conditioned dialogue generation. Pre-training on unlabeled text or dialogue data is not helpful to conditioned generation. Thus, we exploited labeled text data that are easier to collect than labeled dialogues. We expected these data can contribute to better representations of conditions and better use the conditions in natural language generation, which complement what is lacking in the pre-trained models. 

To leverage these two types of data, we proposed a simple and efficient multi-task learning approach. Three tasks are considered: conditioned dialogue generation task on the labeled dialogue data, conditioned language encoding task and conditioned language generation task on the labeled text data. We conducted experiments under persona and topic conditions. Experimental results show that our approach outperforms the state-of-the-art models by leveraging labeled texts, and it also obtains larger improvement in performance comparing to the previous methods leveraging text data.

\bibliography{anthology,acl2020}
\bibliographystyle{acl_natbib}

\clearpage

\appendix

\section{More Implementation Details}
\label{app:details}
For the small-scale datasets, we trained the models until the performances stop to increase on validation set to prevent over-fitting. For large-scale datasets, we fine-tuned the models for four epochs. In Table \ref{tab:runtime}, the average runtime is tested using a 1080Ti GPU device, and the batch size is set to take all of the GPU memories. TA-Seq2Seq and THRED are implemented in TensorFlow. Other models are implemented in PyTorch. Notice that the runtime will be influenced by code implementation in additional to model structure. When experimenting with the small-scale Persona Reddit dataset, we decrease the number of parameters of Sp-Model and MT-Speaker models to 48M and 52M respectively in order to avoid over-fitting. C-Trans-ED loads the pre-training results of GPT. In the original paper, they pre-trained by themselves using a Chinese corpus, which cannot be used in our experiments. 

\begin{table}[h]
\centering
\begin{tabular}{lll}
\hline
\hline
Model & Parameters & Runtime(min/M) \\
\hline 
Sp-Model & 80M & 25 \\
MT-Speaker & 90M & 40 \\
TA-Seq2Seq & 155M & 150 \\
THRED & 174M & 135 \\
C-Trans-ED & 120M & 180 \\
C-Trans-Dec & 126M & 290 \\
BERT & 110M & 140 \\
Ours & 113M & 145\\
\hline
\end{tabular}
\caption{\label{tab:runtime} The number of parameters of each tested approach and the average runtime (minutes) for every million training samples.}
\end{table}

\begin{table}[h]
\centering
\begin{tabular}{ll}
\hline
\hline
Hyper-parameters & Value \\
\hline 
C-Tranformer layers & 2 \\
mask probability & 0.25 \\
max length & 80 \\
batch size & 160 \\
learning rate & 3e-5 \\
warmup proportion & 0.1 \\ 
label smoothing & 0 \\
weight decay & 0.01 \\ 
dropout probability & 0.1 \\
\hline
\end{tabular}
\caption{\label{tab:hyper} Hyper-parameters for our fine-tuning approach. There are in total 6M data. Thus, we use a large batch size. }
\end{table}

Additionally, we explored the idea proposed in \citet{zeng2020open} to decrease finetune-generation discrepancy introduced by MLM training objective. Nevertheless, the conditioned language encoding task cannot employ this method because it applies bi-directional self-attention. Experimental result on small-scale Persona Reddit shows that eliminating this discrepancy helps to decrease perplexity from 55 to 52.

\section{Generation Samples}
\label{app:cases}

\begin{table*}[t]
\centering
\small
\begin{tabular}{ll}
\hline
\hline
\textbf{dialogue His.} & ca n't wait to see the team at full strength when watt and clowney are back . \\
\textbf{Gold Resp.} & i 've been waiting a long time to see watt and clowney healthy together \\
\hline
Sp-Model & i do n't think you know what the word means  \\
MT-Speaker & i do n't think the texans will be able to handle the ball in the first half of the season \\
C-Trans-ED & i do n't know why you 're getting downvoted, i 'm just saying that you 're wrong . \\
C-Trans-Dec & i 'm not sure if i 'd be surprised if they both get hurt . \\
BERT &  i ca n't wait to see watt and clowney at full strength when watt is back . \\
Ours & i do n't think it 's going to happen , but i 'd love to see watt and clowney back . \\
\quad Two-Step FT & i 'd love to see watt and clowney on the same team \\
\quad w/o ctext & i do n't think it 's going to happen , but i 'd love to see watt and clowney on the same team \\
\quad w/o tfidf & i do n't think it 's going to happen , but i 'd love to see watt back .\\
\hline
\hline
Sp-Model & i do n't think he is going to be a good qb .  \\
MT-Speaker & i 'm not sure if this is a good thing .  \\
C-Trans-ED & i 'm not sure what you 're talking about . i do n't know what to say . \\
C-Trans-Dec & i 'm not sure i 'd be surprised if they 're both back  \\
BERT &  i do n't think they 'll be able to do that . \\
Ours & i 'd be so happy to see watt and clowney back .  \\
\quad Two-Step FT & i do n't think that 's what i 'm talking about . \\
\quad w/o ctext & i do n't think they 'll be able to do that . \\
\quad w/o tfidf & i 'm so excited to see watt on the team . \\
\hline
\end{tabular}
\caption{\label{tab:personacase} Responses generated by baselines and our model trained on the large-scale and small-scale Persona Reddit.}
\end{table*}

\begin{table*}[t]
\centering
\small
\begin{tabular}{ll}
\hline
\hline
\textbf{dialogue His.} & what is your favorite food ? [SEP] ice cream . \\
\textbf{Gold Resp.} & what flavor ? \\
\hline
TA-Seq2Seq & what kind of ice cream ? \\
THRED & what kind of ice cream ?  \\
C-Trans-ED & ice cream is the best food i have ever eaten  \\
C-Trans-Dec & i 'm not sure if you 're joking but i 'm not sure if you 're serious .  \\
BERT & i do n't like ice cream . \\
Ours & i do n't like ice cream . it 's coarse and irritating and it gets everywhere . \\
\quad Two-Step FT & i 've never had ice cream . what 's it like ? \\
\quad w/o ctext & i 've never had ice cream . what 's it taste like ? \\
\quad w/o tfidf & what kind of ice cream do you like ? \\
\hline
\hline
C-Trans-ED & ice cream is the best food i 've ever seem . \\
C-Trans-Dec &  i 've never had ice cream . \\
BERT &  i 've never had ice cream . \\
Ours & i do n ' t like ice cream . \\
\quad Two-Step FT &  i like ice cream , but i do n ' t like it . \\
\quad w/o ctext & i 've never had ice cream , but it 's so good . \\
\quad w/o tfidf & i ' ve never had ice cream . \\
\hline
\end{tabular}
\caption{\label{tab:topiccase} Responses generated by baselines and our model trained on the large-scale and small-scale Topic Dialogue. 
}
\end{table*}

\end{document}